\documentclass{article}
\usepackage{spconf,amsmath,graphicx,bm,amsfonts}

\title{COMPARE LEARNING: BI-ATTENTION NETWORK FOR FEW-SHOT LEARNING}
\name{Li Ke \qquad Meng Pan \qquad Weigao Wen \qquad Dong Li}
\address{Alibaba Group, Hangzhou, China}
\begin{document}
%
\maketitle
\begin{abstract}
Learning with few labeled data is a key challenge for visual recognition, as deep neural networks tend to overfit using a few samples only.
One of the Few-shot learning methods called metric learning addresses this challenge by first learning a deep distance metric to determine whether a pair of images belong to the same category, then applying the trained metric to instances from other test set with limited labels.
This method makes the most of the few samples and limits the overfitting effectively.
However, extant metric networks usually employ Linear classifiers or Convolutional neural networks (CNN) that are not precise enough to globally capture the subtle differences between vectors.
In this paper, we propose a novel approach named Bi-attention network to compare the instances,
which can measure the similarity between embeddings of instances precisely, globally and efficiently.
We verify the effectiveness of our model on two benchmarks.
Experiments show that our approach achieved improved accuracy and convergence speed over baseline models.
\end{abstract}
\begin{keywords}
Few-shot, Bi-attention, Compare learning, Metric learning
\end{keywords}
\section{Introduction}
\label{sec:intro}

Deep learning models have achieved great success by using large amounts of labeled data and many iterations to train\cite{he2016deep,huang2017densely,xie2017aggregated}.
This severely limits their scalability to new classes due to annotation cost.
In contrast, learning quickly is a hallmark of human intelligence.
Motivated by the failure of conventional deep learning methods, and inspired by the few-shot learning ability of humans, many recent approaches have made significant progress in few-shot learning\cite{koch2015siamese,vinyals2016matching,snell2017prototypical,finn2017model,sung2018learning,zhang2018metagan}.

Few-shot learning methods can be roughly categorized into three classes: data augmentation, task-based meta learning and compare-based metric learning. 
Data augmentation is a classic technique to increase the amount of available data and just relieved the problem of fewer samples\cite{zhang2018metagan}. 
However, generation models often underperform on few-shot learning tasks\cite{he2016deep}.
An alternative is to merge data from multiple tasks which, however, is not effective due to variances of the data across tasks\cite{wang2018low}.

The task-based meta learning method aims to accumulate experience from learning multiple tasks\cite{naik1992meta,Ravi2016OptimizationAA,hinton1987using,thrun1998learning},
while base-learning focuses on modeling the data distribution of a single task.
A state-of-the art representative of this, namely Model-Agnostic Meta-Learning (MAML), learns to search for the optimal initialization state to fast adapt a base-learner to a new task \cite{finn2017model}. Its task-agnostic property makes it possible to generalize to few-shot supervised learning \cite{finn2018probabilistic,grant2018recasting}. However, in our opinion, there is a limitation of this approach: it usually requires a large number of similar tasks for meta-training which is costly. MAML learns on 240K tasks, and using a shallow CNN makes it can not achieve a good performance. 

The compare methodologies are inspired by humans identifying objects with the help of a comparative way.
Methods include the prototypical networks\cite{snell2017prototypical}, the siamese networks\cite{koch2015siamese} and RelationNets\cite{sung2018learning}.
These approaches focus on learning embeddings that transform the data such that it can be recognised with a fixed nearest-neighbour \cite{snell2017prototypical} or linear \cite{snell2017prototypical,koch2015siamese} classifier.
Among the three, the most related methodology to ours is the RelationNets\cite{sung2018learning} which further defines a deep CNN classifier for metric learning.
Compared to \cite{snell2017prototypical,koch2015siamese}, it can be seen as providing a learnable rather than fixed metric, or non-linear rather than linear classifier.

\begin{figure*}[t]

\begin{minipage}[b]{1.0\linewidth}
  \centering
  \centerline{\includegraphics[width=15cm]{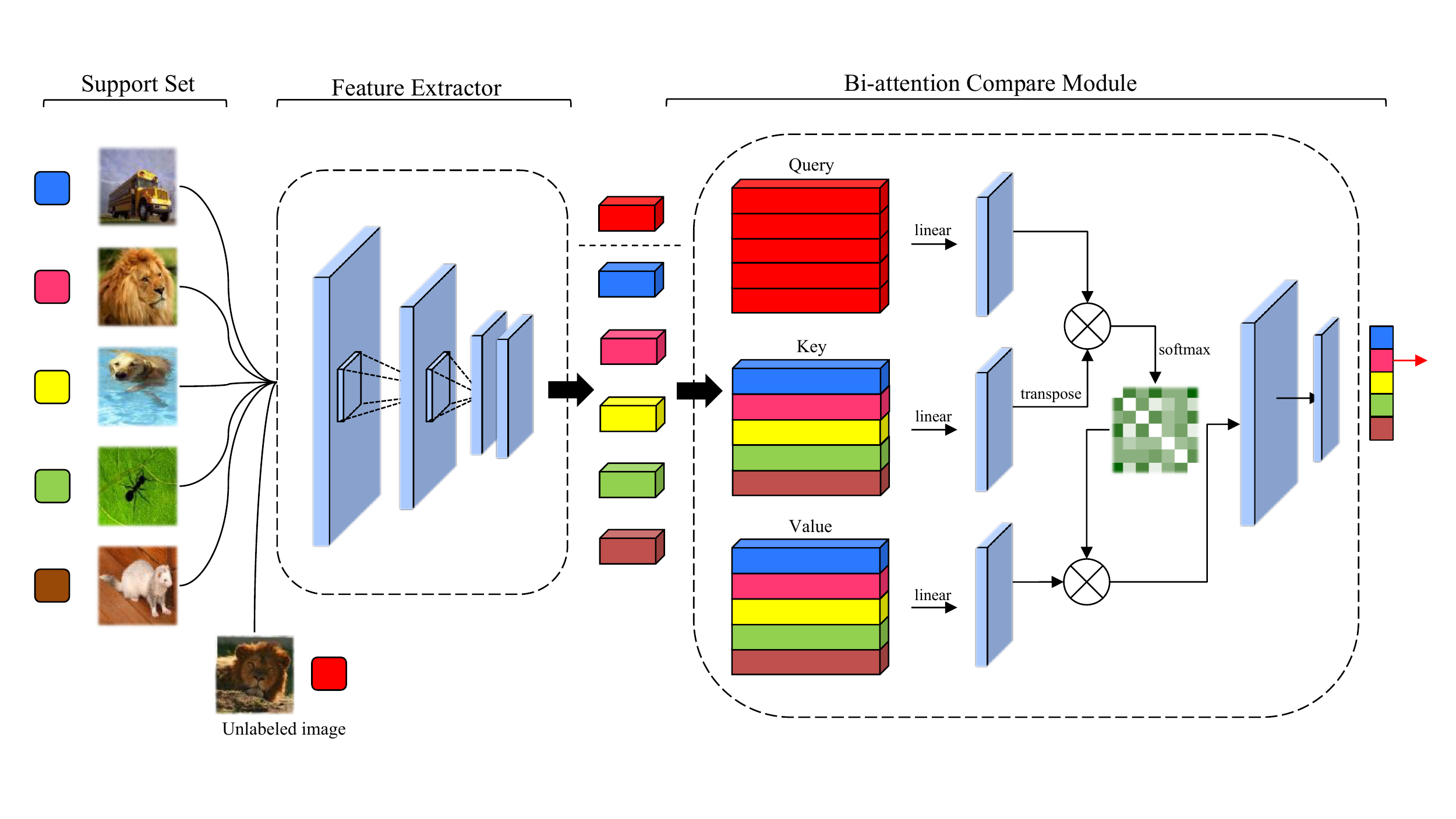}}
\end{minipage}
\caption{The overview of Bi-Attention Network for a 5-way 1-shot image recognition task. The $\bigotimes$ denotes matrix multiplication.}
\label{fig:fig1}
\end{figure*}

But when using the CNN network as a comparison module, its feature extraction is based on sliding kernels, leading it focuses more on comparing local information.
As a result, RelationNets\cite{sung2018learning} is unable to globally learn the most fine-grained features for comparison from the embedding vector.
Regarding the issue above, our framework further defines a new Compare Network named Bi-attention Network. We are inspired by the transformer architecture \cite{vaswani2017attention,devlin2018bert} that has an outstanding performance, especially in the NLP field.
The core of that is the multi-heads self-attention mechanism, which is a triplet in the form of (query, key and value).
It first matches a query with all the keys, calculates the weight of each key for the query, and finally weighted sum the values corresponding to each key to obtain the output.
Unlike self-attention, we reform it into bi-attention to calculate the relationship score between the unlabeled query and a small number of labeled samples.
Compared to RelationNets \cite{sung2018learning} we benefit from an element-wise and position-independent attention strategy, thus all elements from the two embeddings will be compared to each other entirely with the finest granularity. 
To the best of our knowledge, our approach is the only direct use of attention mechanisms as metrics, rather than as a component to enhance other network structures.
Overall our contribution is to provide a better comparing method that elegantly solves the few-shot learning problem.

We evaluate models on two publicly available benchmark datasets: miniImageNet\cite{vinyals2016matching} 
and CIFAR100\cite{krizhevsky2009learning,oreshkin2018tadam}.
Experiment results demonstrate that our model outperforms almost all extant compare networks and most of the other few-shot methods. 
In addition, the convergence speed comparison also demonstrates strong evidence that our model improves training efficiency significantly. 

We organize this paper as follows. In section 2 we define the problem and describe our proposed methods in detail. In section 3 we show our experimental settings and compare experiment results with baselines. We summarize this paper in section 4.

\section{Methods}
\label{sec:format}

In this section, we describe our proposed method in detail. Model architecture is depicted in Figure 1.

\subsection{Description on FewShot Learning}
\label{ssec:description}

Few-shot learning models are trained on a labeled dataset ${D}_{train}$, and tested on ${D}_{test}$ with only a few labeled samples per class.
The class sets are disjoint between ${D}_{train}$ and ${D}_{test}$.
Successful approaches rely on an episodic training process: sampling small samples from the labeled dataset ${D}_{train}$ during each training episode that mimics the few-shot learning setting of test tasks.
In general, each episode $e$ is created by first sampling $N$ categories from the ${D}_{train}$ and then sampling two sets of images from these categories:
(1) the support set $S_e={(s_i,y_i)}_{i=1}^{N \times K}$ containing $K$ examples for each of the $N$ categories, thus called $N$-way $K$-shot classification problem.
(2) the query set $Q_e={(q_j,y_j)}_{j=1}^{M}$ containing $M$ different examples from the same $N$ categories.

During each training episode for a few-shot classification, minimizing the loss of the prediction on samples in query set, given the support set.
Then, in the test phase, we sample the support and query set from ${D}_{test}$ as the same as ${D}_{train}$, but the label of query set is unseen, needs to be predicted.

\subsection{Model Structure}
\label{ssec:subhead}
Bi-attention compare network consists of two modules: a feature extractor $f_\theta$ (e.g. convolutional layers in ResNets \cite{he2016deep}) and a compare module $f_\varphi$.

In episode $e$, All $s$ and $q$ firstly feed into feature extractor $f_\theta$ by batch to get the embedding.
Let $s^k_n\in S_e$, where $k=1,2,...,K$, $n=1,2,...,N$, and $N,K$ keeps the same as section \ref{ssec:description}.
Feed $s^k_n$ into $f_\theta$, we have $f_\theta(s_n^k)\in\mathbb{R}^{l \times d \times d}$, where $l$, $d$ denote the output number of channel and spatial size. 
In order to get class vectors of $N$ different classes from support sets, we element-wise sum the embeddings of the $K$ samples in each category $n$ to form the class embedding:
\begin{equation}
\label{equation1}
    c_n = \sum\limits_{k=1}^{K}f_\theta(s_k^n),  c_n\in\mathbb{R}^{l \times d \times d}
\end{equation}
Then, let $C=[c_1,c_2,...,c_N]\in\mathbb{R}^{N \times l \times d \times d}$ denote the class vector in a task. 
About query samples, let $q_j\in Q_e$, where $j=1,2,...,M$, $M$ is the number of query samples.
We combine them into the query vector $Q$. Let $p_j=f_\theta(q_j)$, then $Q=[p_1,p_2,...,p_M]\in\mathbb{R}^{M \times l \times d \times d}$.
Finally, we respectively repeat $C$ by $M$ times and $Q$ by $N$ times, which will be able to compare each query sample with all class embeddings.
We reshape them and feed them into $f_\varphi$ to get the compare score:
\begin{equation}
\label{equation1}
    X = f_\varphi(Q,C) \quad Q,C \in\mathbb{R}^{MN \times l_h \times d_{h}}
\end{equation}
Where $X\in\mathbb{R}^{M \times N}$, and $x_{j,n}$ is the comparing score between $q_j$ and $c_n$. That is, $X=\{x_{j,n}\mid j=1,...,M, n=1,...,N\}$.
Besides, $d_{h}$ is a hyperparameter of hidden size, and $l_h$ could be obtained by $l_h=\frac{l \times d^2}{d_{h}}$ accordingly.

\subsection{Bi-attention}
\label{ssec:subhead}
In this section, we describe the details about our bi-attention compare module, which is in the previous chapter annotated as $f_\varphi$. The main idea of bi-attention is that giving a query $q_j$, firstly match it against the key of each class by inner product and softmax operator to get a bi-attention matrix, which could be considered as the degree of matching between query and key of each class. Finally, the compare score of $q_j$ with classes is then returned as the sum of all the class embeddings weighted by bi-attention matrix, defined as follow:
\begin{equation}
    BiAttn(Q,C)=softmax(\frac{QC^T}{\sqrt{d_z}}){C}
\end{equation}
Where $d_z$ is a scale factor. 
In our approch, we apply a Multi-Head Bi-Attention structure to enhance the compare ability of model.
\begin{equation}
    head_i=BiAttn(QW_i^Q,CW_i^C)
\end{equation}
\begin{equation}
    H=Concat(head_1,head_2,...,head_h)W^O
\end{equation}
Where linearly mapping parameter matrices: $W_i^Q\in\mathbb{R}^{d_{h} \times d_q}$, $W_i^C\in\mathbb{R}^{d_{h} \times d_c}$ and $W^O\in\mathbb{R}^{hd_c \times d_{h}}$. $h$ is the number of head. For each head, we use $d_q=d_c=d_{h}/h$.
Finally, we use two layers of linear mapping to further summarize the comparison results.
\begin{equation}
    f_\varphi(Q,C)=sigmoid(HW_1+b_1)W_2+b_2
\end{equation}
Where $W_1\in\mathbb{R}^{{d_h} \times 1}$ and $W_2\in\mathbb{R}^{{l_h} \times 1}$ are still linearly mapping parameters, $b_1$ and $b_2$ are bias.
\subsection{Training Objective}
\label{ssec:subhead}
The train objective is performing the SGD update to minimize the Cross-Entropy loss during each epoch:
\begin{equation}
    \theta,\varphi = \arg \min \limits_{\theta,\varphi}\sum\limits_{j=1}^{M}\sum\limits_{n=1}^{N}Loss(x_{j,n},y_j)
\end{equation}
\begin{equation}
    Loss(x_{j,n},y_j) = -(y_j==n)logx_{j,n}
\end{equation}
where $y_j$ is the ground truth class label of $q_j$ and $n$ is the class which $q_j$ compares to. $x_{j,n}$ is comparing score returned by equation \ref{equation1}. $\theta,\varphi$ contains all the learnable parameters of the model.

\section{EXPERIMENTS}
\label{sec:experiments}
In this section, we evaluate the proposed Bi-attention network in
terms of few-shot recognition accuracy and model convergence
speed. 

\subsection{Datasets}
\label{ssec:datasets}
We conduct few-shot learning classification tasks on two benchmarks, miniImageNet and CIFAR100.
\begin{itemize}
	\item \textbf{miniImageNet}. miniImageNet \cite{vinyals2016matching} was widely used in related works. In total, there are 100 classes with 600 samples per class. 
	These 100 classes are divided into 64, 16, and 20 classes respectively for sampling tasks for training, 
	validation and test, following related works.
	\item \textbf{CIFAR100}. CIFAR100 was proposed in \cite{krizhevsky2009learning}. Same as miniImageNet, we divide it that contains 100 object classes with 600 images for each class. The 100 classes belong to 20 super-classes, we use leaf class as the label \cite{oreshkin2018tadam}. 
Training data are from 60 classes belonging to 12 super-classes,
validation and test sets contain 20 classes belonging to 4 super-classes, respectively.
These splits accord to super-classes, thus minimize the information overlap between training and val/test tasks.
So it is more challenging in terms of lower image resolution and stricter training-test splits than miniImageNet.
\end{itemize}

\begin{figure*}[t]
\centering
\begin{minipage}[b]{.24\linewidth}
  \centering
  \centerline{\includegraphics[width=4.2cm]{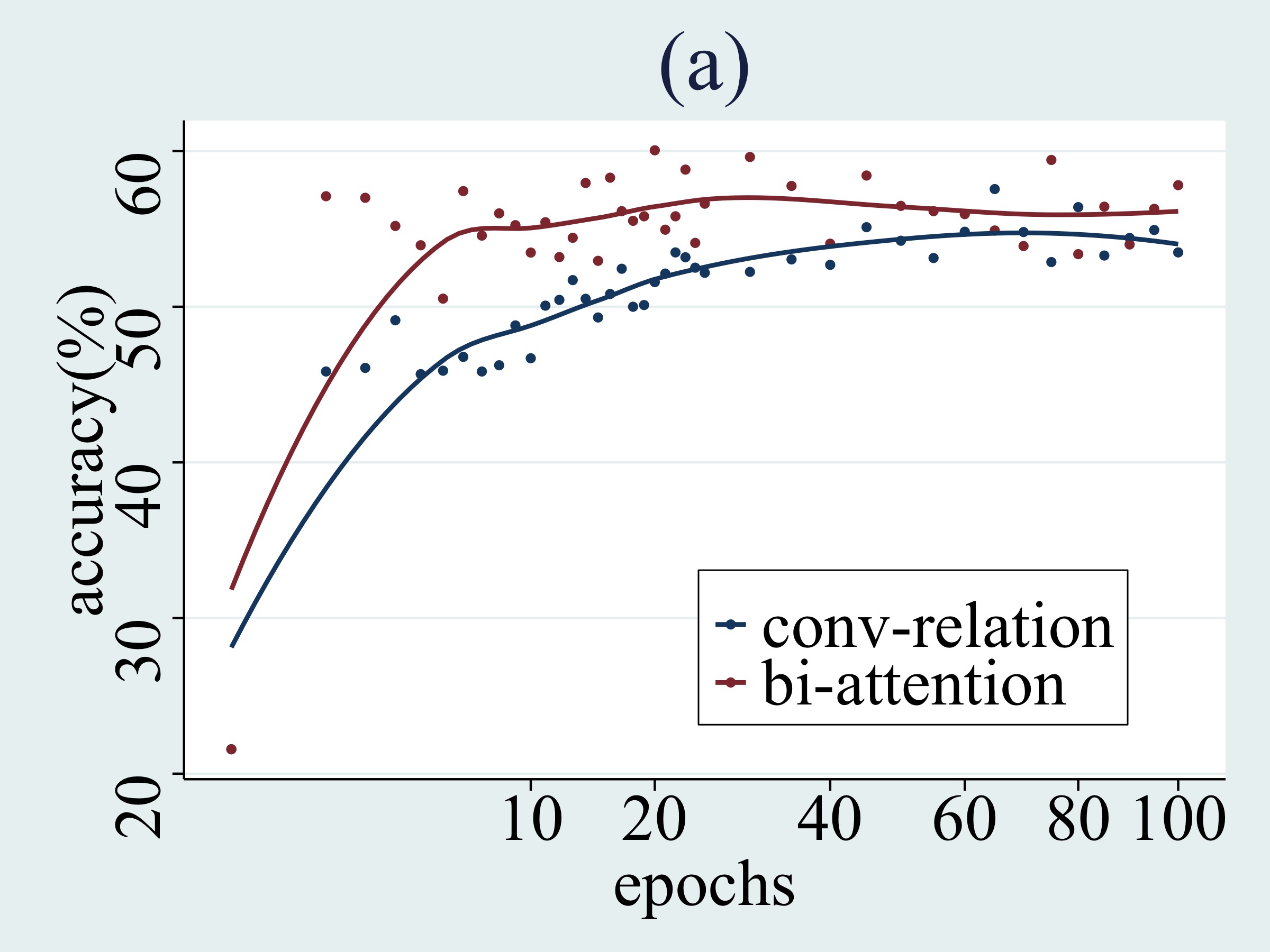}}
\end{minipage}
\begin{minipage}[b]{.24\linewidth}
  \centering
  \centerline{\includegraphics[width=4.2cm]{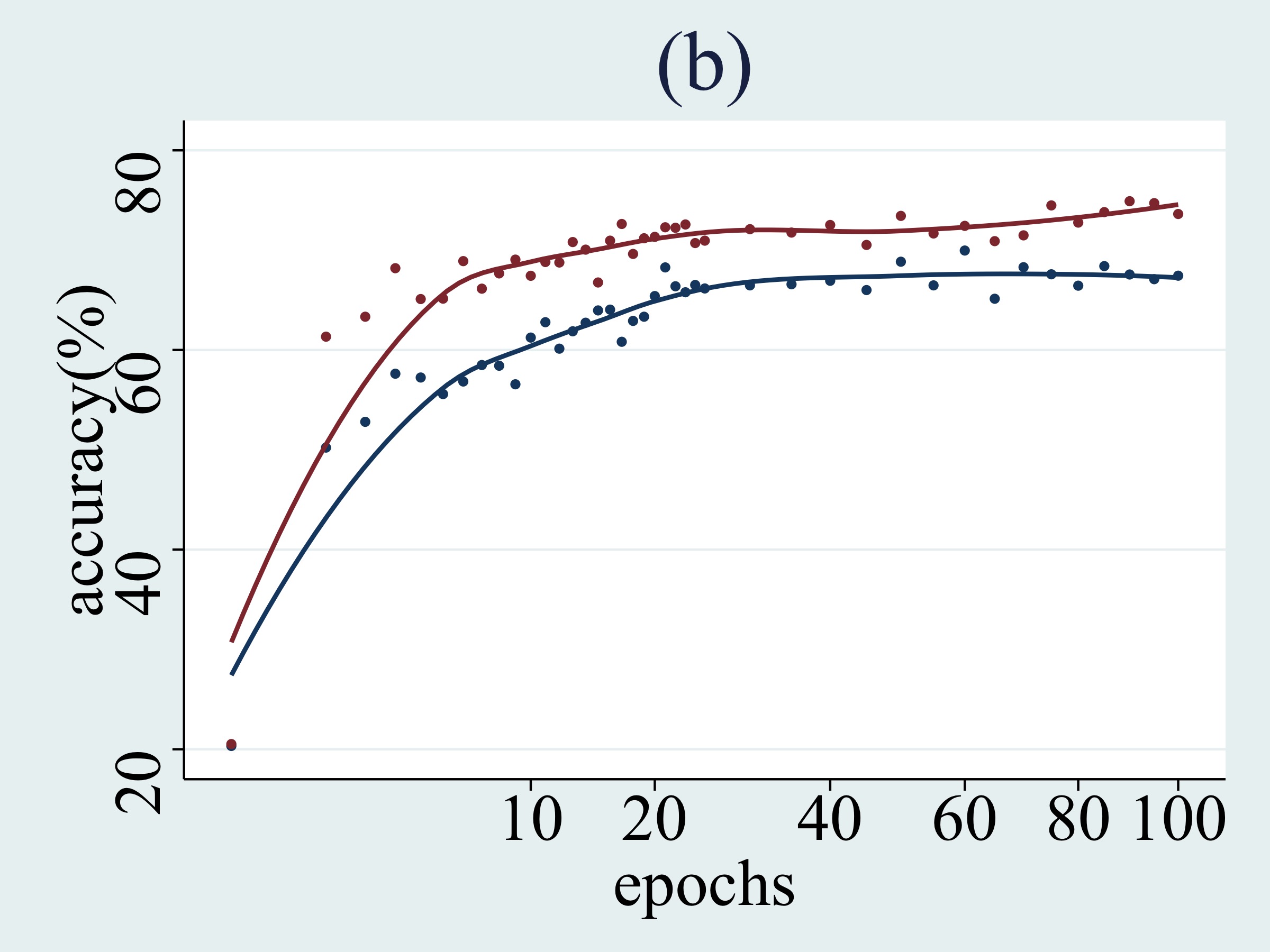}}
\end{minipage}
\begin{minipage}[b]{0.24\linewidth}
  \centering
  \centerline{\includegraphics[width=4.2cm]{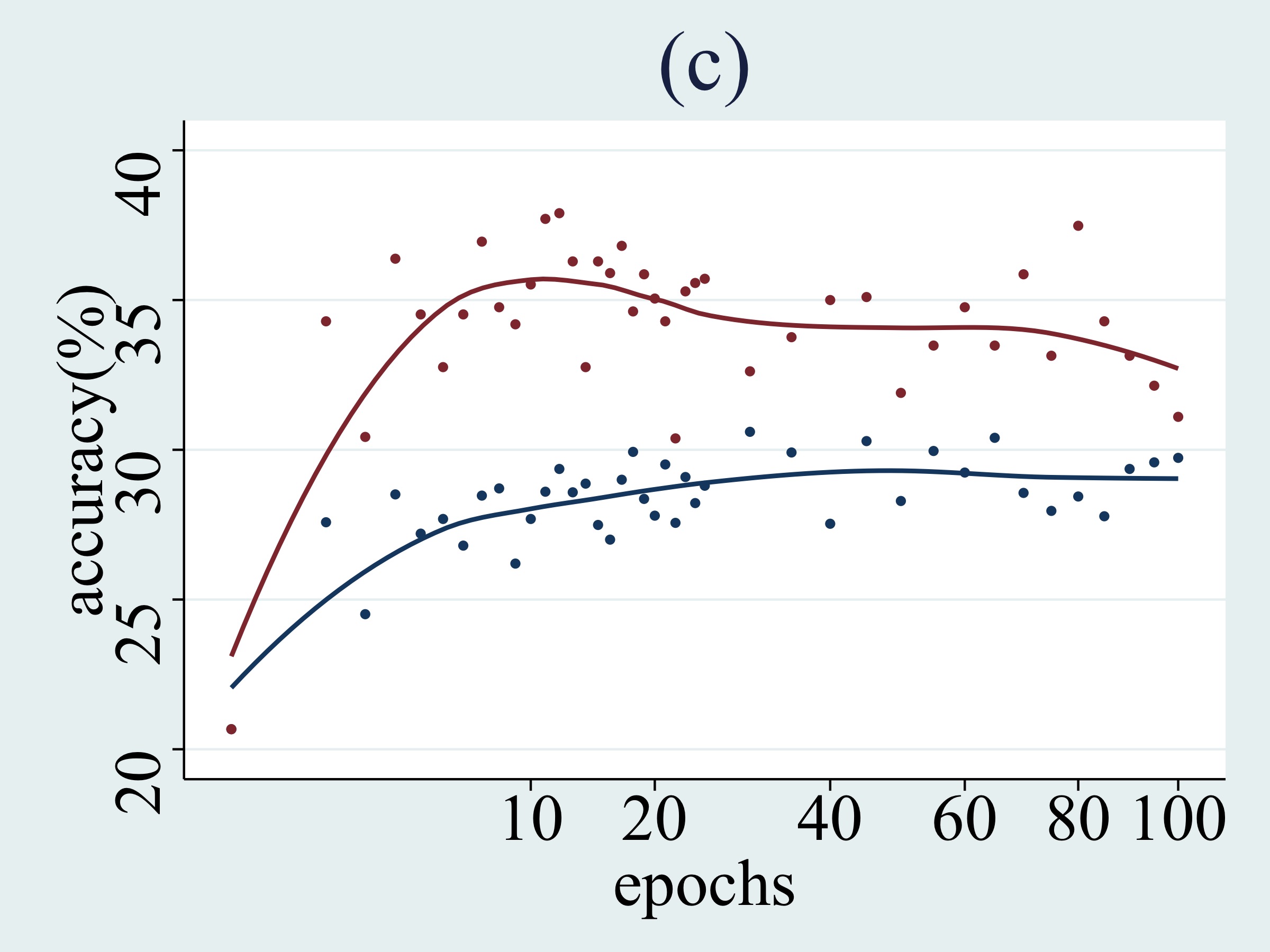}}
\end{minipage}
\begin{minipage}[b]{0.24\linewidth}
  \centering
  \centerline{\includegraphics[width=4.2cm]{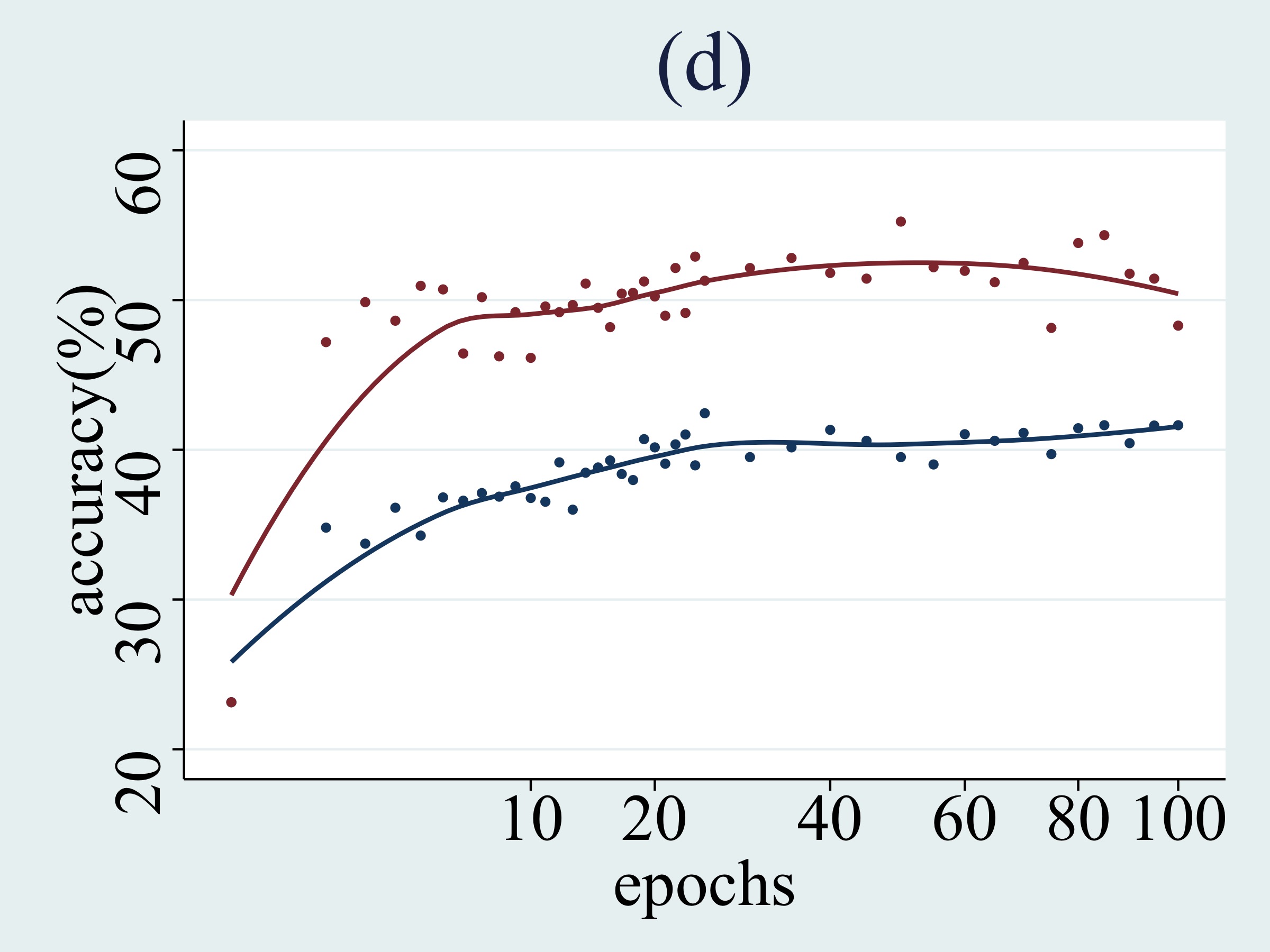}}
\end{minipage}
\caption{(a)(b) show the results of 1-shot and 5-shot on miniImageNet; (c)(d) show the results of 1-shot and 5-shot on CIFAR100.}
\label{fig:fig2}
\end{figure*}

\subsection{Experimental Settings}
\label{ssec:settings}
About feature extractor $f_\theta$, we apply a classic ResNet-12 network \cite{he2016deep,franceschi2018bilevel} to extract embedding of images. It contains 4 residual blocks and each block has 3 CONV layers
with $3\times3$ kernels. At the end of each residual block, a
$2\times2$ max-pooling layer is applied.
And initialize it on a multi-classification task only with images from the train set.
Regarding compare module $f_\varphi$, we use 8 heads of bi-attention with the hidden size of 128, that is $h = 8$ and $d_h = 128$.
We evaluate models with 1-shot-5-way and 5-shot-5-way settings sharing the same hyperparameter.
The learning rate starts at $0.001$, will be halved every $10$ epoch.
Model trains within $120$ epoch and each epoch included $100$ tasks of comparative learning. Besides, we sample 600 random tasks respectively for validation and test.


\subsection{Baseline Models}
\label{ssec:baseline}
We compare our model with the following models. Covers different genres of representative methods in Few-shot learning.

\begin{itemize}
	\item \textbf{MAML}: proposed by Chelsea Finn et al. \cite{finn2017model} Learns to search for the optimal initialization state to fast adapt to a new few-shot task. It is model-agnostic.
	\item \textbf{MetaGAN}: by introducing an adversarial generator conditioned on tasks, Zhang et al. \cite{zhang2018metagan} proposed the method that augments vanilla few-shot methods with the ability to learn sharper decision boundary.
	\item \textbf{ProtoNets}: Prototypical networks presented by Snell et al. \cite{snell2017prototypical}
 learn a metric space in which classification can be performed by computing distances to prototype representations of each class.
	\item \textbf{RelationNets}: Flood Sung et al. \cite{sung2018learning} proposed a relation network that learns to learn a deep CNN metric to compare support image and the query image.
\end{itemize}

\subsection{Experiment Results}
\label{ssec:result}
Following previous work, all accuracy results are averaged over 600 test tasks and are reported with 95\% confidence intervals.
As reported in Table \ref{tab:table1}, our model has got the top performance in both 1-shot and 5-shot settings with miniImageNet dataset.
Especially in 5-shot learning, we significantly improve accuracy by 3\% than MetaGAN.
About the CIFAR100 dataset in Table \ref{tab:table2}, although ProtoNets has always performed well, our method performed slightly better in the 5-shot setting. Furthermore, our method performed significantly in the 1-shot setting.
Statistics demonstrate that our model can be able to among the top models on all the baselines. 
Overall, our method surpasses other comparing networks proving that our Bi-attention element-wise compare structure is better than the CNN or Linear comparison structure.

Besides, we evaluate our training efficiency, as shown in figure \ref{fig:fig2}.
The reason why we compare RelationNets\cite{sung2018learning} with our approach is that both methods in Table \ref{tab:table1} are composed of deep metric learning for few-shot tasks,
causing them comparable. 
The baseline model RelationNets was reimplemented by the same setting in section \ref{ssec:settings} using the source code of \cite{sung2018learning} for fairness.
The first 10 epochs clearly showed that our method can achieve faster convergence in both data sets and two tasks.
Experimental results showed that our concise Bi-attention network is more favorable as a deep metric learning module in comparison tasks than multi-layer convolution networks. Both convergence speed and accuracy are better than the latter.

\begin{table}[h!]
  \begin{center}
    \caption{The 5-way, 1-shot and 5-shot classification testing accuracy(\%) on miniImageNet dataset.}
    \label{tab:table1}
    \begin{tabular}{c|c|c} 
      \hline
      \textbf{Few-shot method} & \textbf{1 shot} & \textbf{5 shot}\\
      \hline
      MAML & $48.70\pm1.84$ & $63.11\pm0.92$ \\
      ProtoNets & $49.42\pm0.78$ & $68.20\pm0.66$\\
      RelationNets & $50.44\pm0.82$ & $65.32\pm0.70$\\
      MetaGAN & $52.71\pm0.64$ & $68.63\pm0.67$\\
      Our Approach & \bm{$53.74$}$\pm$\bm{$0.89$} & \bm{$71.90$}$\pm$\bm{$0.76$}\\
      \hline
    \end{tabular}
  \end{center}
\end{table}
\begin{table}[h!]
  \begin{center}
    \caption{The 5-way, 1-shot and 5-shot classification testing accuracy(\%) on CIFAR100 dataset. MetaGAN has released neither source code nor result of CIFAR100 till now, so it doesn't appear below.}
    \label{tab:table2}
    \begin{tabular}{c|c|c} 
      \hline
      \textbf{Few-shot method} & \textbf{1 shot} & \textbf{5 shot}\\
      \hline
      MAML & $38.10\pm1.70$ & $50.40\pm0.99$ \\
      ProtoNets & $36.70\pm0.68$ & $56.50\pm0.71$\\
      RelationNets & $36.56\pm0.70$ & $48.86\pm0.65$\\
      Our Approach & \bm{$39.08$}$\pm$\bm{$0.81$} & \bm{$56.89$}$\pm$\bm{$0.79$}\\
      \hline
    \end{tabular}
  \end{center}
\end{table}

\section{CONCLUSION}
\label{sec:experiments}
In this work, we proposed a novel metric learning method: Bi-attention Network which further globally compares the similarities between images in detail with an element-wise way,
that helps to address the few-shot learning classification issue.
We conducted extensive experiments on two challenging benchmarks. 
Experiment results demonstrate that our model outperforms most of extant models, we even achieve the best accuracy on two datasets under part of settings. 
Besides, our model achieves high efficiency for learning.
Overall, we believe that our approach can be applied to address another few-shot learning task by incurring limited training costs.


\vfill
\pagebreak

\bibliographystyle{IEEEbib}
\bibliography{strings,refs}

\end{document}